\documentclass[10pt]{cai26}

\usepackage{bm}
\usepackage{bbm}
\usepackage{algorithm}
\usepackage{algpseudocode}
\usepackage{subcaption}
\usepackage{makecell,rotating}
\setcellgapes{5pt}
\usepackage{diagbox}
\usepackage{hyperref}
\usepackage{caption}
\usepackage{makecell}
\usepackage[none]{hyphenat}

\begin{document}
\def\conferenceyear{2026}
\volumeheader{39}{0}
\begin{center}

\title{Diagnosing and Repairing Factual Errors in RAG under Budget Constraints}
\maketitle

\thispagestyle{empty}
\pagenumbering{gobble}

\begin{tabular}{cc}
Soroush Hashemifar\upstairs{\affilone}, Havva Alizadeh Noughabi\upstairs{\affiltwo}, Fattane Zarrinkalam\upstairs{\affilone,\affiltwo,*}, Ali Dehghantanha\upstairs{\affiltwo}
\\[0.25ex]
{\small \upstairs{\affilone} College of Engineering, University of Guelph, Guelph, ON, Canada} \\
{\small \upstairs{\affiltwo} Cyber Science Lab, School of Computer Science, University of Guelph, Guelph, ON, Canada} \\
\end{tabular}

\emails{
  \upstairs{*} fzarrink@uoguelph.ca 
}
\vspace*{0.1in}
\end{center}

\begin{abstract}
Retrieval-Augmented Generation (RAG) improves the factuality of large language models by grounding responses in external evidence, yet real-world deployments remain fragile. Failures often stem from missing or weakly relevant evidence, as well as from generation that does not faithfully reflect the retrieved context. Many existing approaches rely on fine-tuning, privileged access to internal model signals, or resource-insensitive escalation strategies, which limits their practicality in black-box and budget-constrained settings. We propose \textsc{D2R-RAG} (Diagnose-to-Repair RAG), a model-agnostic and resource-aware framework that combines lightweight failure diagnosis with adaptive repair. \textsc{D2R-RAG} derives interpretable failure signatures from observable signals in the query, retrieved evidence, and generated response, and then selects from a small set of corrective actions under explicit latency and VRAM constraints. Experiments on FEVER and HotpotQA show that \textsc{D2R-RAG} improves reliability over recent baselines and achieves better accuracy--efficiency trade-offs across multiple compute budgets. The code is available at \url{https://github.com/CyberScienceLab/D2R-RAG/}.
\end{abstract}

\copyrightnotice

\section{Introduction}
\sloppy 
Retrieval-Augmented Generation (RAG)~\cite{lewis2020retrieval} has become a standard approach for improving the factuality of large language model (LLM) outputs by grounding generation in external evidence rather than relying solely on parametric memory. In many real deployments, however, RAG remains unreliable: correct answers may fail when the retriever misses key evidence, when retrieved passages are relevant but incomplete, or when the generator produces content that is not supported by the provided context~\cite{xie2024adaptive}. These failure modes are stochastic and non-uniform, and they are most problematic in settings with strict latency and hardware budgets (e.g., limited GPU memory, rate-limited APIs, or cost-constrained edge/cloud deployments), where iterative retrieval or resource-intensive reranking are computationally prohibitive~\cite{ray2025metis}.

Recent work has proposed dynamic and self-correcting RAG variants from three directions: (1) training-time modifications, (2) inference-time control, and (3) learning-based decision policies. Self-RAG~\cite{asai2024self} trains generators to emit reflection control tokens that decide when to retrieve and how to critique intermediate outputs, improving factuality through self-assessment, but requires fine-tuning and architectural coupling that limits use in black-box or API-based settings. Moreover, DRAGIN~\cite{su2403dragin} uses token-level uncertainty and other internal dynamics to trigger retrieval during inference, offering lightweight, training-free control, yet relies on logit access (unavailable in closed-source LLM APIs). Finally, learning-based policies like MBA-RAG~\cite{tang2025mba} (multi-armed bandit balancing quality and efficiency, dependent on an external query-complexity predictor) and QueryBandits~\cite{cho2025querybandits} (bandit-based query rewriting using semantic and lexical signals) demonstrate that online decision-making can improve robustness, but typically target specific pipeline stages or require additional supervision or modeling components. 

This paper addresses \emph{model-agnostic, resource-aware RAG recovery in black-box settings}, arguing that reliable recovery requires two capabilities: a \emph{lightweight diagnostic} distinguishing retrieval-side evidence insufficiency from generation-side unfaithfulness using only observable artifacts, and the \emph{least-cost corrective action} under explicit latency and VRAM budgets. We operationalize these principles in \textsc{D2R-RAG} (Diagnose-to-Repair RAG), which performs triangulated verification by combining (1) textual entailment checks between retrieved evidence and both the query and response, with (2) structured consistency checks that align relational triples extracted from the response against a knowledge graph to identify entity/predicate mismatches. These signals yield an interpretable failure signature separating missing evidence from unsupported generation. Repair selection is then formulated as a contextual multi-armed bandit~\cite{langford2007epoch} where each arm corresponds to a specific intervention (query rewriting, retrieval depth or mode adjustment, cross-encoder reranking, or index refresh), and a LinUCB policy~\cite{li2010contextual} learns to balance expected factual recovery against computational cost. Notably, the framework requires neither generator fine-tuning nor access to internal logits, enabling deployment with closed-source LLM APIs. 

We evaluate \textsc{D2R-RAG} on fact verification and multi-hop question answering using FEVER~\cite{Thorne18Fever} and HotpotQA~\cite{yang2018hotpotqa}. Our evaluation emphasizes both factual performance and operational efficiency under multiple latency/VRAM budget regimes. Beyond aggregate results, we analyze how the learned policy allocates high-cost interventions across different failure signatures and how performance changes as resource budgets tighten.


\section{Methodology}\label{sec:proposed}

\begin{figure}[t]
    \centering
    \includegraphics[scale=0.5]{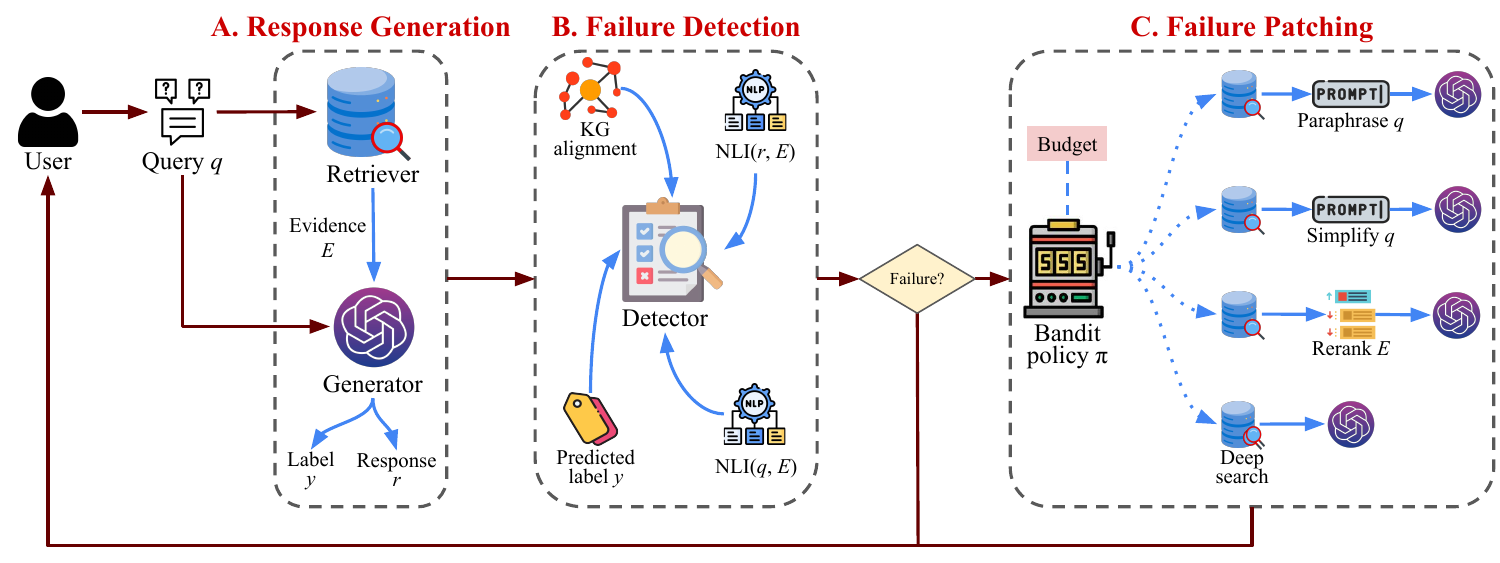}
    \caption{Overview of \textsc{D2R-RAG}.}
    \label{fig:architecture}
\end{figure}

Figure~\ref{fig:architecture} overviews the \textsc{D2R-RAG}, which augments a RAG pipeline with two modules.

\textbf{(1) Failure Diagnosis.} 
This stage determines whether the initial RAG output is trustworthy and identifies the likely failure type, using query $q$, evidence $E$, response $r$, and predicted label $y$, which is suitable for black-box deployment. The goal is to produce an interpretable failure signature that separates retrieval-side insufficiency from generation-side unfaithfulness, rather than perfectly explaining every error. We derive the signature using three complementary signals: \textit{query entailment} $e^q$, \textit{response entailment} $e^r$, and \textit{KG alignment status} $\kappa$~\cite{naveen2023nli}. See Appendix~\ref{apx:diagnosis} for additional details.

Given $(\kappa, e^q, e^r, y)$, our failure types are: wrong-predicate (WP) for KG conflict, insufficient-evidence (IE) when retrieved evidence does not entail the query ($e^q=\textsf{neutral}$), wrong-response (WR) when the response is not entailed by retrieved evidence ($e^r\in\{\textsf{neutral},\textsf{contradict}\}$), and label--evidence mismatch (LEM) for FEVER when the predicted $y$ conflicts the query-context entailment. For FEVER, IE and LEM cases emit the abstention label \emph{Unverified} rather than a potentially correct label for the wrong reason, improving trustworthiness and prioritizing retrieval-oriented repairs (evidence-gated label prediction).

\textbf{(2) Adaptive Repair via Contextual Bandits.} 
\textsc{D2R-RAG} performs single-shot repair by selecting a patch and re-running the RAG pipeline, as a contextual multi-armed bandit (CMAB)~\cite{langford2007epoch}: the learner observes context vector $\mathbf{x}\in\mathbb{R}^d$, chooses action $a\in\mathcal{A}$, and receives a reward reflecting output quality and resource cost.
The action space $\mathcal{A}$ is intentionally small and black-box deployable: (1) prompt-level rewrites (paraphrasing and simplification) to reduce query ambiguity~\cite{cho2025querybandits}; (2) reranker activation using a cross-encoder to improve evidence precision; and (3) retrieval-level interventions adjusting depth $k$, switching between BM25 and dense retrieval, or refreshing the index. 
We learn the repair policy with LinUCB~\cite{li2010contextual}, which assumes a linear reward model $\hat{r}(a\mid\mathbf{x})=\mathbf{x}^\top \boldsymbol{\theta}_a$ and selects actions by maximizing an upper-confidence bound (more details in Appendix~\ref{apx:context}):

\begin{equation}
    a^*=\arg\max_{a\in\mathcal{A}}
    \left(
        \mathbf{x}^\top \boldsymbol{\theta}_a
        + \alpha\sqrt{\mathbf{x}^\top \mathbf{A}_a^{-1}\mathbf{x}}
    \right).
    \label{eq:argmax}
\end{equation}
where $\alpha$ controls exploration and $\mathbf{A}_a$ is the action-specific covariance matrix. After applying $a^*$, the revised output and measured action cost define the observed reward.
We use a resource-aware reward scoring repairs by output reliability and budget compliance. Let $L_a$ and $V_a$ denote the latency and additional VRAM usage incurred by action $a$, with per-action budgets $B_L$ and $B_V$. The reward is:
\begin{equation}
    r(a) = \frac{1}{4}\left( \mathbbm{1}_{NF} + \mathbbm{1}_{KG} + 2\cdot \mathbbm{1}_{NLI} \right)
    \prod_{x \in \{L, V\}} \mathbbm{1}\!\left[x_{a} \leq B_x\right]\!\left( 1 - \frac{x_{a}}{B_x} \right).
    \label{eq:reward}
\end{equation}
where $\mathbbm{1}_{NF}$ indicates a \textsf{NoFailure} diagnosis after repair, $\mathbbm{1}_{KG}$ indicates triple-level consistency, and $\mathbbm{1}_{NLI}$ indicates entailment support. Hard gates $\mathbbm{1}[x_a\le B_x]$ zero out rewards for budget-violating actions, while multiplicative terms encourage cheaper actions among feasible ones. Following each interaction, LinUCB updates only the selected action:
\begin{equation}
    \mathbf{A}_{a} \leftarrow \mathbf{A}_{a} + \mathbf{x}\mathbf{x}^\top,\qquad
    \mathbf{b}_{a} \leftarrow \mathbf{b}_{a} + r(a)\mathbf{x},\qquad
    \boldsymbol{\theta}_{a} \leftarrow \mathbf{A}_{a}^{-1}\mathbf{b}_{a}.
    \label{eq:linucb_update}
\end{equation}

\section{Experiments}\label{sec:experiments}

We investigate three research questions: 
\textbf{RQ1:} Does \textsc{D2R-RAG} improve factual quality while respecting deployment costs compared to baselines? \textbf{RQ2:} Does the policy adapt interventions to specific failure modes rather than using a uniform strategy? \textbf{RQ3:} How does the reward formulation in \textsc{D2R-RAG} determine the learned repair strategy, and how stable is this under varying latency/VRAM budgets?

\subsection{Experimental Setup}\label{subsec:expsetup}

\textbf{Datasets.} 
We evaluate \textsc{D2R-RAG} on FEVER~\cite{Thorne18Fever} (2{,}000 claim verification instances) and HotpotQA~\cite{yang2018hotpotqa} (1{,}000 multi-hop question answering instances).

\textbf{Baselines.}
We evaluate \textsc{D2R-RAG} against \textit{Naive-RAG} (a single-pass RAG), \textit{Query Paraphrase}~\cite{deng2023rephrase} (rewriting the query conditioned on the retrieved context), \textit{Context Expansion} (increasing dense retrieval depth from $k$ to $k' = 20$), and \textit{Thompson Sampling}~\cite{Thompson1933thompson} (replacing LinUCB with a context-free policy).
All methods utilize a \textit{GPT-4o-mini} generator and 128-token chunks with 16-token overlap, within LlamaIndex.

\textbf{Evaluation Metrics.}
We report \textit{Exact Match (EM)} for HotpotQA, \textit{Accuracy (ACC)} for FEVER, and response \textit{Relevance} and \textit{Faithfulness}. Efficiency is measured by Latency (retrieval to final output) and peak VRAM (maximum GPU memory during execution). 

\textbf{Implementation Settings.} 
We use \textit{DeBERTa-v3-large-nli} for entailment check and \textit{Babelscape/rebel-large} for triple extraction. We impose 3s latency and 6 MB VRAM per query. The repair policy is trained online for two epochs, with exploration parameter $\alpha{=}2$.

\subsection{Results}

\textbf{RQ1.}
Table~\ref{tab:overallresults} shows that \textsc{D2R-RAG} improves factual quality while maintaining efficiency constraints: bold and underlined values represent best and second-best results per dataset/metric, respectively. On FEVER, it improves ACC from 56.3\% (\textit{Naive-RAG}) to 61.5\% (\textit{Thompson Sampling}) and 60.8\% (\textit{LinUCB}), with \textit{Thompson Sampling} achieving the best cost profile (1.14s, 0.17 MB) compared to \textit{LinUCB} (1.47s, 0.36 MB). On HotpotQA, \textit{Thompson Sampling} reaches 40.4\% EM with 71.52\% faithfulness, while \textit{LinUCB} matches 39.8\% EM but at higher latency (3.41s vs 2.37s); \textit{LinUCB} occasionally exceeds the VRAM budget (3.41 MB), since budget constraints are enforced per individual repair action rather than end-to-end. Overall, \textit{LinUCB} prioritizes quality at higher resource cost, while \textit{Thompson Sampling} offers a stronger quality--efficiency trade-off (see Appendix~\ref{apx:results}).

\begin{table}[htbp]
    \centering
    \begin{tabular}{l l c c c c c c}
        \toprule
        Dataset & Method & Latency & VRAM & Relevance & Faithfulness & ACC & EM \\
        \midrule

        \multirow{5}{*}{FEVER} 
        & Naive-RAG & \underline{1.16} & \underline{0.20} & 50.16 & 91.67 & 56.3 & - \\
        & Query Paraphrase & 1.32 & \underline{0.20} & 50.09 & \underline{92.69} & 56.7 & - \\
        & Context Expansion & 1.35 & \underline{0.20} & \underline{50.24} & \textbf{92.82} & \textbf{61.8} & - \\
        & \textsc{D2R-RAG} (LinUCB) & 1.47 & 0.36 & \textbf{50.80} & 92.34 & 60.8 & - \\
        & \textsc{D2R-RAG} (T.Sampling) & \textbf{1.14} & \textbf{0.17} & 50.00 & 92.37 & \underline{61.5} & - \\
        \midrule

        \multirow{5}{*}{HotpotQA}
        & Naive-RAG & \textbf{1.52} & \textbf{0.36} & 28.39 & 68.58 & - & 36.2 \\
        & Query Paraphrase & \underline{1.60} & \textbf{0.36} & 27.90 & \textbf{74.65} & - & 39.8 \\
        & Context Expansion & \textbf{1.52} & \textbf{0.36} & \textbf{32.14} & 70.70 & - & \textbf{40.6} \\
        & \textsc{D2R-RAG} (LinUCB) & 3.41 & \underline{1.04} & \underline{31.66} & 69.89 & - & 39.8 \\
        & \textsc{D2R-RAG} (T.Sampling) & 2.37 & 1.23 & 30.69 & \underline{71.52} & - & \underline{40.4} \\
        \bottomrule
    \end{tabular}
    \caption{Comparison of D2R-RAG and baselines on FEVER and HotpotQA.}
    \label{tab:overallresults}
\end{table}

\begin{figure}[htbp]
    \centering
    \begin{subfigure}[t]{0.5\textwidth}
        \centering
        \includegraphics[height=1.4in]{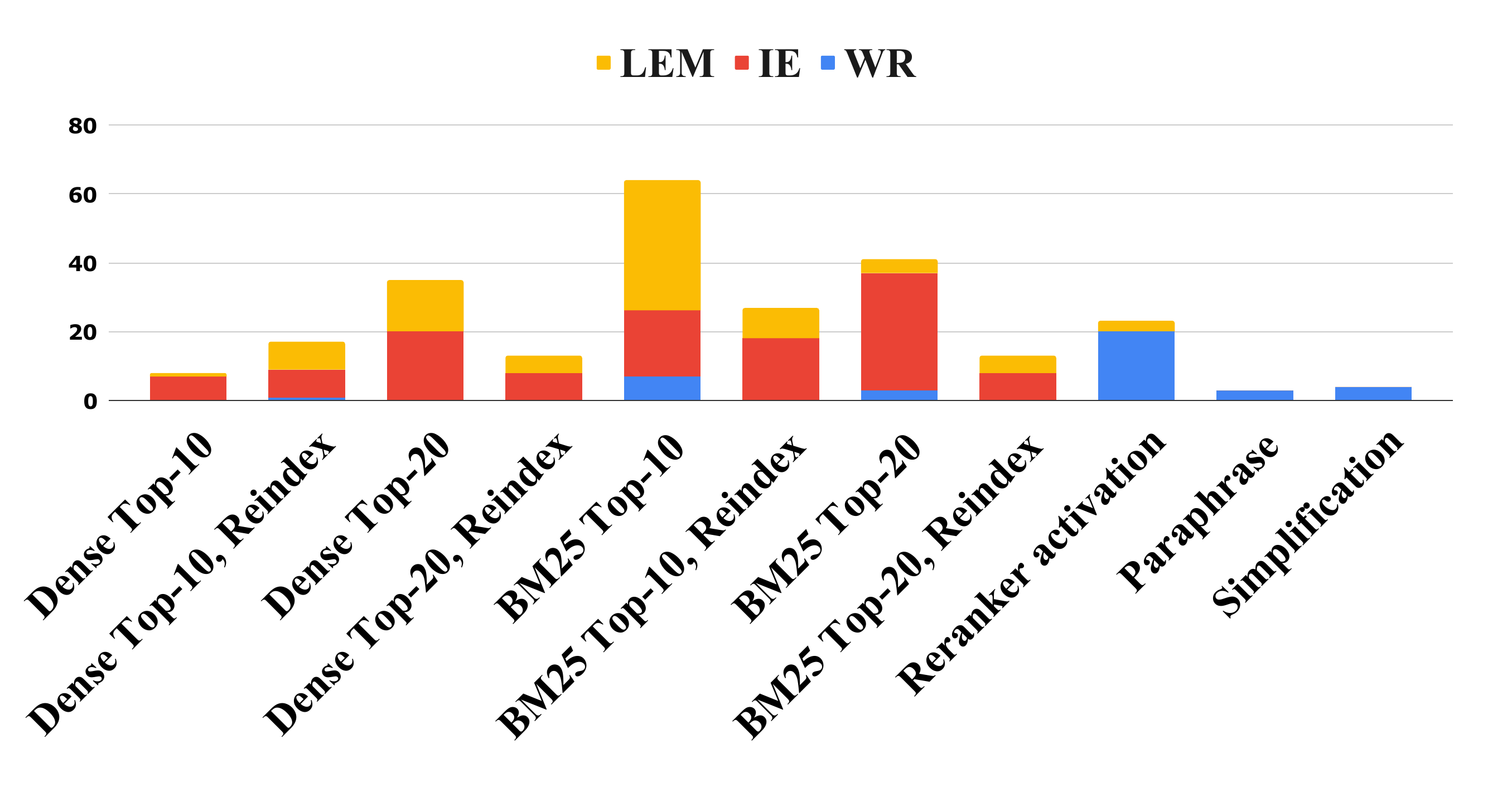}
        \caption{LinUCB on FEVER}
    \end{subfigure}%
    ~ 
    \begin{subfigure}[t]{0.5\textwidth}
        \centering
        \includegraphics[height=1.4in]{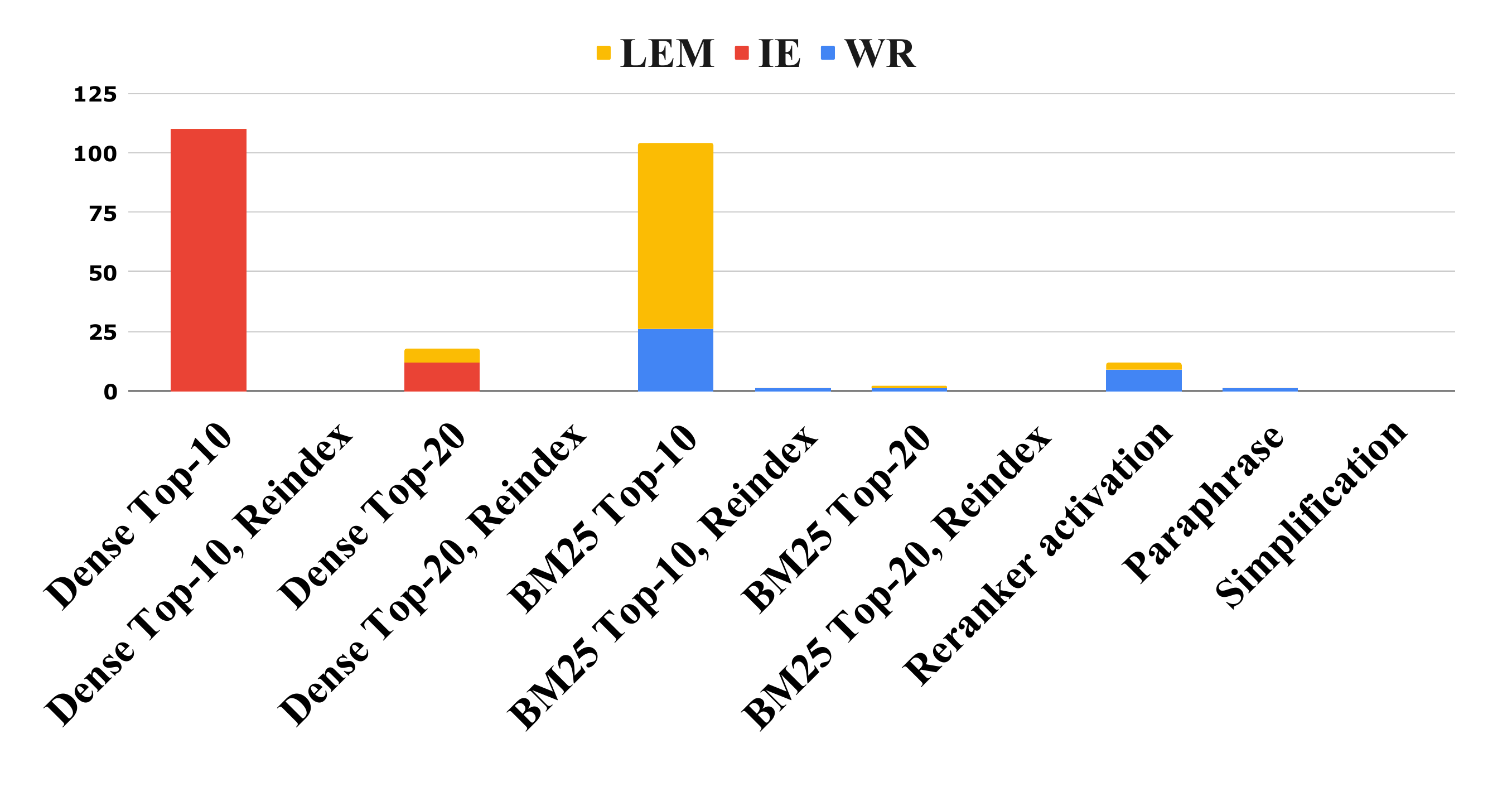}
        \caption{Thompson Sampling on FEVER}
    \end{subfigure}
    \\
    \begin{subfigure}[t]{0.5\textwidth}
        \centering
        \includegraphics[height=1.4in]{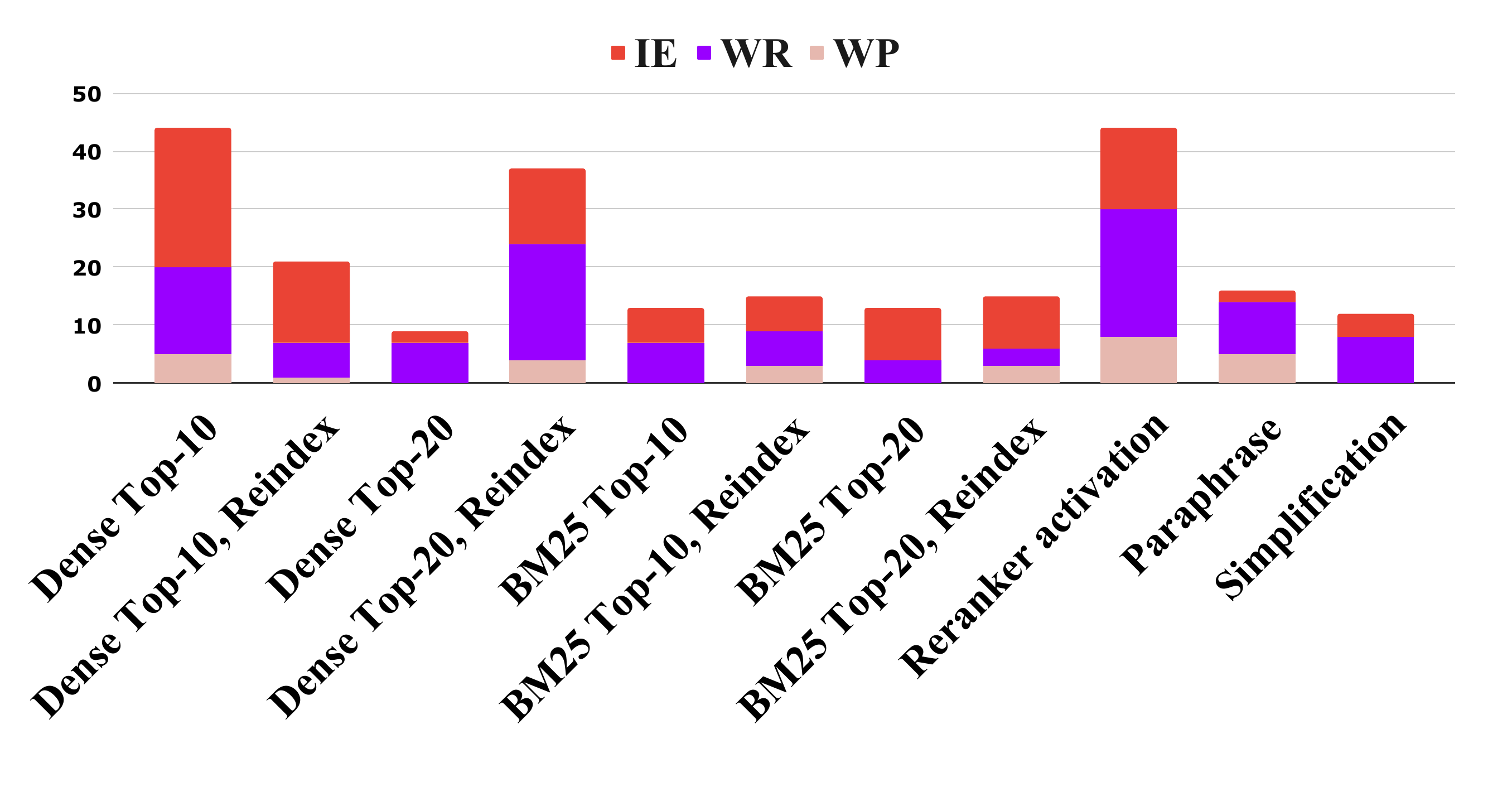}
        \caption{LinUCB on HotpotQA}
    \end{subfigure}
    ~ 
    \begin{subfigure}[t]{0.5\textwidth}
        \centering
        \includegraphics[height=1.4in]{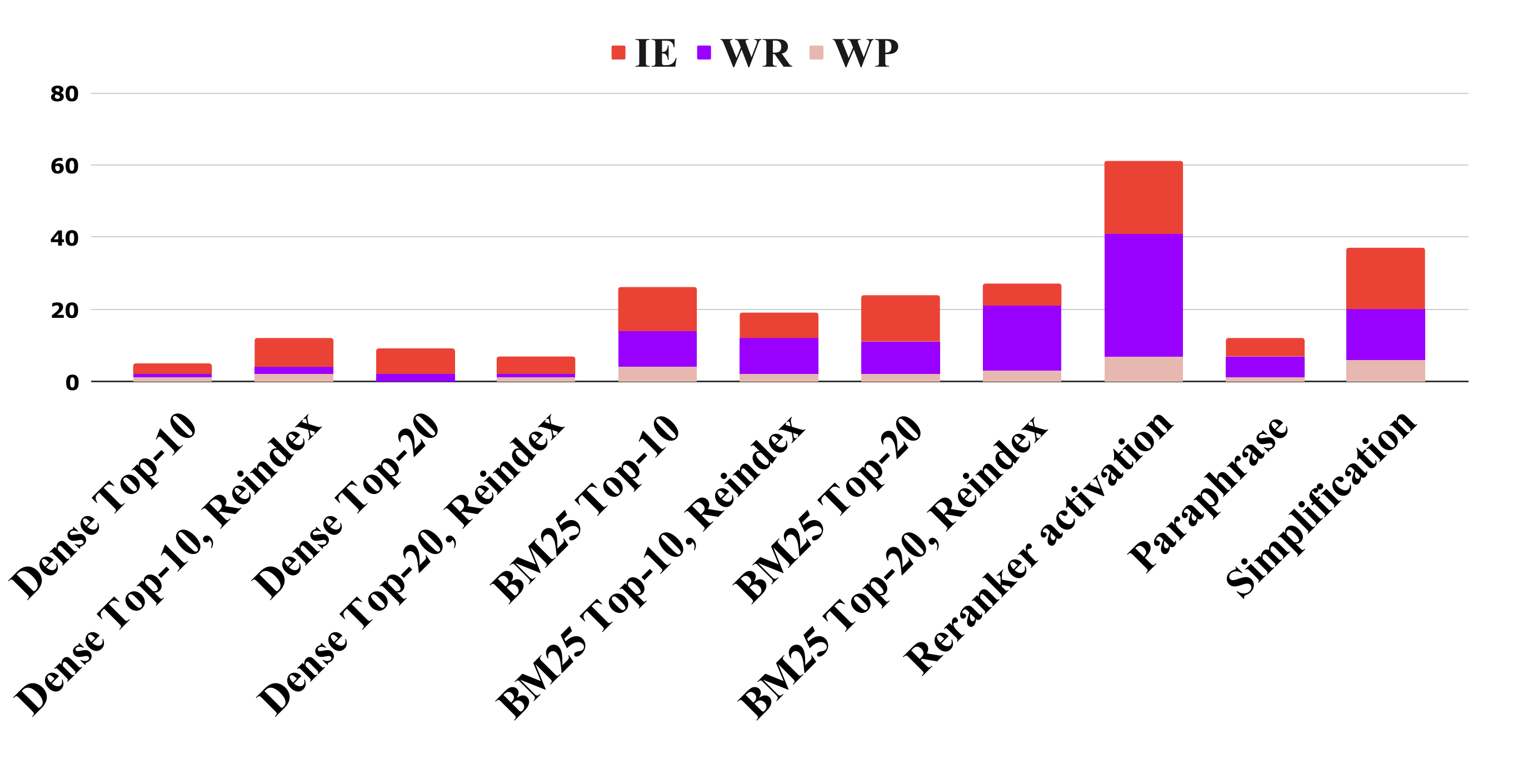}
        \caption{Thompson Sampling on HotpotQA}
    \end{subfigure}
    \caption{Analysis of failure type frequency exposed by \textsc{D2R-RAG}.}
    \label{fig:flabelfeq}
\end{figure}

\textbf{RQ2.}
Figure~\ref{fig:flabelfeq} shows that \text{D2R-RAG} conditions actions on failure type. On FEVER, IE cases favor retrieval escalation ($\sim$60--80\% on deeper dense/BM25), while on HotpotQA, WR and WP failures route to reranker activation, indicating that selecting the right passages (not merely retrieving more) is critical for multi-hop questions. 
\textit{LinUCB} explores broader action distribution, while \textit{Thompson Sampling} exploits more (concentrates on high-utility repairs), yet both preserve the same failure-to-action mapping.

\textbf{RQ3.}
Table~\ref{tab:overallresultsreward} ablates the two mechanisms that encode deployment constraints: \textit{Unweighted}, which removes the soft resource-weighting terms, and \textit{Unconstrained}, which removes the hard budget gates. On HotpotQA, \textit{Unconstrained} reduces latency (3.41$\rightarrow$2.03s) and VRAM (1.04$\rightarrow$0.84 MB) with minimal EM drop, while \textit{Unweighted} improves quality metrics (Relevance 31.66$\rightarrow$32.20\%; Faithfulness 69.89$\rightarrow$70.59\%) with near-identical EM. On FEVER, both ablations improve ACC (60.8$\rightarrow$61.3-61.4\%) and reduce memory. 
Budget sensitivity tests (Table~\ref{tab:overallresultsbudget}) show \textit{Stringent} (0.7$\times$) cuts VRAM on both datasets while preserving performance, and \textit{Relaxed} (1.5$\times$) lowers latency and boosts faithfulness. Notably, \text{D2R-RAG} achieves the highest HotpotQA EM at highest cost, demonstrating that budget shaping controls quality--cost trade-offs. More results are provided in Appendix~\ref{apx:results}.

\begin{table}[htbp]
    \centering
    \fontsize{7.5}{8.5}\selectfont
    \setlength{\tabcolsep}{3pt}
    \renewcommand{\arraystretch}{1.0}
    \begin{tabular}{l l cccccc}
        \toprule
        Dataset & Variant & Latency & VRAM & Relevance & Faithfulness & ACC & EM \\
        \midrule
        \multirow{3}{*}{FEVER} 
        & Unconstrained & \textbf{1.40} & \underline{0.27} & \textbf{50.91} & \underline{92.61} & \underline{61.3} & -- \\
        & Unweighted & \underline{1.46} & \textbf{0.19} & 50.64 & \textbf{92.62} & \textbf{61.4} & -- \\
        & \textsc{D2R-RAG} & 1.47 & 0.36 & \underline{50.80} & 92.34 & 60.8 & -- \\
        \midrule
        \multirow{3}{*}{HotpotQA} 
        & Unconstrained & \textbf{2.03} & \underline{0.84} & 30.36 & 69.46 & -- & 39.0 \\
        & Unweighted & \underline{2.79} & \textbf{0.68} & \textbf{32.20} & \textbf{70.59} & -- & \underline{39.4} \\
        & \textsc{D2R-RAG} & 3.41 & 1.04 & \underline{31.66} & \underline{69.89} & -- & \textbf{39.8} \\
        \bottomrule
    \end{tabular}
    \caption{Overall performance comparison of reward variants.}
    \label{tab:overallresultsreward}
\end{table}

\begin{table}[htbp]
    \centering
    \fontsize{7.5}{8.5}\selectfont
    \setlength{\tabcolsep}{3pt}
    \renewcommand{\arraystretch}{1.0}
    \begin{tabular}{l l cccccc}
        \toprule
        Dataset & Variant & Latency & VRAM & Relevance & Faithfulness & ACC & EM \\
        \midrule
        \multirow{3}{*}{FEVER} 
        & Stringent & 2.22 & \textbf{0.24} & \textbf{51.01} & \textbf{92.80} & \underline{60.6} & - \\
        & Relaxed & \textbf{1.42} & \underline{0.32} & 50.63 & \underline{92.76} & \underline{60.6} & - \\
        & \textsc{D2R-RAG} & \underline{1.47} & 0.36 & \underline{50.80} & 92.34 & \textbf{60.8} & - \\
        \midrule
        \multirow{3}{*}{HotpotQA} 
        & Stringent & \textbf{2.23} & \textbf{0.66} & 29.99 & \underline{71.46} & - & 39.2 \\
        & Relaxed & \underline{2.31} & \underline{0.72} & \underline{30.81} & \textbf{71.57} & - & \underline{39.4} \\
        & \textsc{D2R-RAG} & 3.41 & 1.04 & \textbf{31.66} & 69.89 & - & \textbf{39.8} \\
        \bottomrule
    \end{tabular}
    \caption{Overall performance comparison across budget variants.}
    \label{tab:overallresultsbudget}
\end{table}

\section{Conclusion}\label{sec:conclusion}

We introduced \textsc{D2R-RAG}, a repair framework for black-box RAG systems with lightweight diagnostics and adaptive patch decisions. 
Experiments suggest that reliability improves most when repair is targeted to specific failure signatures.
Limitations include: (1) diagnostic signal imperfections, since KG alignment and NLI models can struggle with noisy evidence and multi-hop reasoning, cascading errors into suboptimal repairs; and (2) the policy is constrained to exposed repair actions, limiting its capability or overusing costly patches when cheaper alternatives are unavailable. 

\section*{GenAI Usage Disclosure}
OpenAI’s ChatGPT was used to refine sentence clarity and grammatical correctness during manuscript preparation.

\section*{Acknowledgments}
This work was supported in part by the NSERC-CSE Research Community Grants (ALLRP 598786-24), NSERC Canada Research Chair Grant (CRC-2024-00017), and the National Cybersecurity Consortium (2025-1601) projects. Researchers funded through the NSERC-CSE Research Communities Grants do not represent the Communications Security Establishment Canada or the Government of Canada. Any research, opinions or positions they produce as part of this initiative do not represent the official views of the Government of Canada.

\printbibliography[heading=subbibintoc]

\newpage
\appendix

\section{Diagnostic Signals}\label{apx:diagnosis}

We derive failure signatures from three complementary diagnostic signals.
The first and second are semantic support via textual entailment: an NLI model evaluates whether retrieved passages address the query and whether the response is supported by evidence. Concretely, each passage in $E$ is treated as a premise and the query (or response) as a hypothesis; passage-level predictions are aggregated into two coarse entailment states $e^q$ and $e^r$. 
Intuitively, $e^q$ captures whether retrieved evidence is relevant and sufficient for the information need, while $e^r$ captures whether the generated output remains faithful to the evidence.
The third source is structured consistency derived from relational triples. Entailment signals can miss fine-grained relational errors, e.g., response mentions the correct entities but asserts an incorrect predicate or swaps roles. To capture such errors, we extract relations from the response and check whether the stated relations align with the underlying knowledge source. 
In practice, this signal is most useful for distinguishing “topically plausible but relationally wrong” answers from answers that are merely missing evidence.
Algorithm~\ref{alg:failuredetector} summarizes the rule-based diagnosis.

\section{Context Representation and Training Process}\label{apx:context}

To form the context vector $\mathbf{x}$ used by the repair policy, in our implementation, we concatenate: (1) a semantic representation of the query, (2) the diagnosed failure type, (3) the entailment and triple-alignment statuses, and (4) normalized remaining latency and VRAM budgets. This representation supports cost-sensitive choices: evidence-insufficiency signatures favor retrieval-side escalation, while evidence-present but unsupported-response signatures prioritize higher-precision evidence selection (e.g., reranking) or query rewrites. Algorithm~\ref{alg:train} summarizes training procedure. The process is one-shot: for each diagnosed failure, the policy selects a single action and performs exactly one additional RAG pass.

\begin{figure}[htbp]
    \centering
    \begin{minipage}{0.49\textwidth}
        \begin{algorithm}[H]
            \caption{\textsc{Failure Diagnosis}}
            \label{alg:failuredetector}
            \footnotesize
            \begin{algorithmic}[1]
                \Require query $q$, evidence set $E$, response $r$, predicted task label $y$
                \Ensure failure type $f \in \{\textsf{NoFail}, \textsf{WP}, \textsf{IE}, \textsf{WR}, \textsf{LEM}\}$
                
                \State $\kappa \gets \Call{TripleAlignStatus}{r}$
                \State $e^{q} \gets \Call{AggregateNLI}{E, q}$
                \State $e^{r} \gets \Call{AggregateNLI}{E, r}$
                
                \If{$\kappa = \textsf{conflict}$}
                    \State \Return $\textsf{WP}$
                \EndIf
                \If{$e^{q} = \textsf{neutral}$}
                    \State \Return $\textsf{IE}$
                \EndIf
                \If{$e^{r} \neq \textsf{entail}$}
                    \State \Return $\textsf{WR}$
                \EndIf
                \If{$y$ is provided \textbf{and} $\Call{LabelCompatible}{y, e^{q}} = \textsf{false}$}
                    \State \Return $\textsf{LEM}$
                \EndIf
                
                \State \Return $\textsf{NoFailure}$
            \end{algorithmic}
        \end{algorithm}
    \end{minipage}
    \hfill 
    \begin{minipage}{0.49\textwidth}
        \begin{algorithm}[H]
        \caption{Bandit Policy Learning}
        \label{alg:train}
        \footnotesize
        \begin{algorithmic}[1]
                \Require dataset $\mathcal{D}$, default RAG configuration $P$
                \Ensure learned bandit parameters $\{\mathbf{A}_a,\mathbf{b}_a\}_{a\in\mathcal{A}}$ (equivalently $\{\boldsymbol{\theta}_a\}_{a\in\mathcal{A}}$)
                \State Initialize $\mathbf{A}_a \gets \mathbf{I}$, $\mathbf{b}_a \gets \mathbf{0}$ for all $a\in\mathcal{A}$
                \For{each epoch}
                    \For{each query $q \in \mathcal{D}$}
                        \State $(r,y,E) \gets \Call{RAG}{q,P}$
                        \State $f \gets \Call{DetectFailure}{q,E,r,y}$
                        \If{$f \neq \textsf{NoFailure}$}
                            \State $\mathbf{x} \gets \Call{BuildContext}{q,f,E,r}$
                            \State $a \gets \Call{SelectAction}{\mathbf{x}, \{\mathbf{A}_a,\mathbf{b}_a\}}$
                            \State $P' \gets \Call{ApplyPatch}{P,a}$
                            \State $(r',y',E') \gets \Call{RAG}{q,P'}$
                            \State $f' \gets \Call{DetectFailure}{q,E',r',y'}$
                            \State $r(a) \gets \Call{ComputeReward}{f',L_a,V_a}$
                            \State \Call{UpdateLinUCB}{$a,\mathbf{x},r(a)$}
                        \EndIf
                    \EndFor
                \EndFor
                \State \Return $\{\mathbf{A}_a,\mathbf{b}_a\}_{a\in\mathcal{A}}$
            \end{algorithmic}
        \end{algorithm}
    \end{minipage}
\end{figure}

\section{Diagnostic Analysis and Budget Sensitivity}\label{apx:results}

Regarding factual quality, Table~\ref{tab:diagnosticresults} shows that correct predictions concentrate in entailment-supportive regimes: on HotpotQA, both policies yield 170+ correct answers under response-entailment versus far fewer under contradiction, while on FEVER, most correct predictions fall under entailed response with contradiction states remaining rare and mostly incorrect. Frequent missing KG alignment indicates triple extraction falls short for many queries, yet policies still succeed when entailment signals provide adequate support. 

\begin{table}[htbp]
    \centering
    \begin{tabular}{l l cc cc}
        \toprule
        \multirow{2}{*}{Diagnostic} & \multirow{2}{*}{Status} & \multicolumn{2}{c}{HotpotQA} & \multicolumn{2}{c}{FEVER} \\
        \cmidrule(lr){3-4}\cmidrule(l){5-6}
        & & LinUCB & T.Sampling & LinUCB & T.Sampling \\
        \midrule
        \multirow{4}{*}{KG Alignment} & Consistent & 80 {\small (65)} & 73 {\small (60)} & 112 {\small (81)} & 111 {\small (75)} \\
        & Conflict & 9 {\small (16)} & 8 {\small (19)} & 0 {\small (0)} & 0 {\small (0)} \\
        & Missing & 110 {\small (220)} & 121 {\small (219)} & 494 {\small (309)} & 502 {\small (308)} \\
        & No Triplet & 0 {\small (0)} & 0 {\small (0)} & 2 {\small (2)} & 2 {\small (2)} \\
        \cmidrule(lr){1-6}
        \multirow{3}{*}{Query Entailment} & Entail & 115 {\small (148)} & 112 {\small (155)} & 350 {\small (203)} & 351 {\small (195)} \\
        & Contradict & 59 {\small (107)} & 57 {\small (95)} & 258 {\small (139)} & 264 {\small (140)} \\
        & Neutral & 25 {\small (46)} & 33 {\small (48)} & 0 {\small (50)} & 0 {\small (50)} \\
        \cmidrule(lr){1-6}
        \multirow{3}{*}{Resp. Entailment} & Entail & 173 {\small (229)} & 174 {\small (230)} & 600 {\small (377)} & 611 {\small (375)} \\
        & Contradict & 7 {\small (30)} & 6 {\small (22)} & 8 {\small (10)} & 4 {\small (9)} \\
        & Neutral & 19 {\small (42)} & 22 {\small (46)} & 0 {\small (5)} & 0 {\small (1)} \\
        \bottomrule
    \end{tabular}
    \caption{Diagnostic signal counts. Each cell reports the \# correct (incorrect) instances, where correctness means EM{=}1 on HotpotQA and label accuracy (ACC){=}1 on FEVER.}
    \label{tab:diagnosticresults}
\end{table}

\begin{figure}[htbp]
    \centering
    \begin{subfigure}[t]{0.5\textwidth}
        \centering
        \includegraphics[height=1.5in]{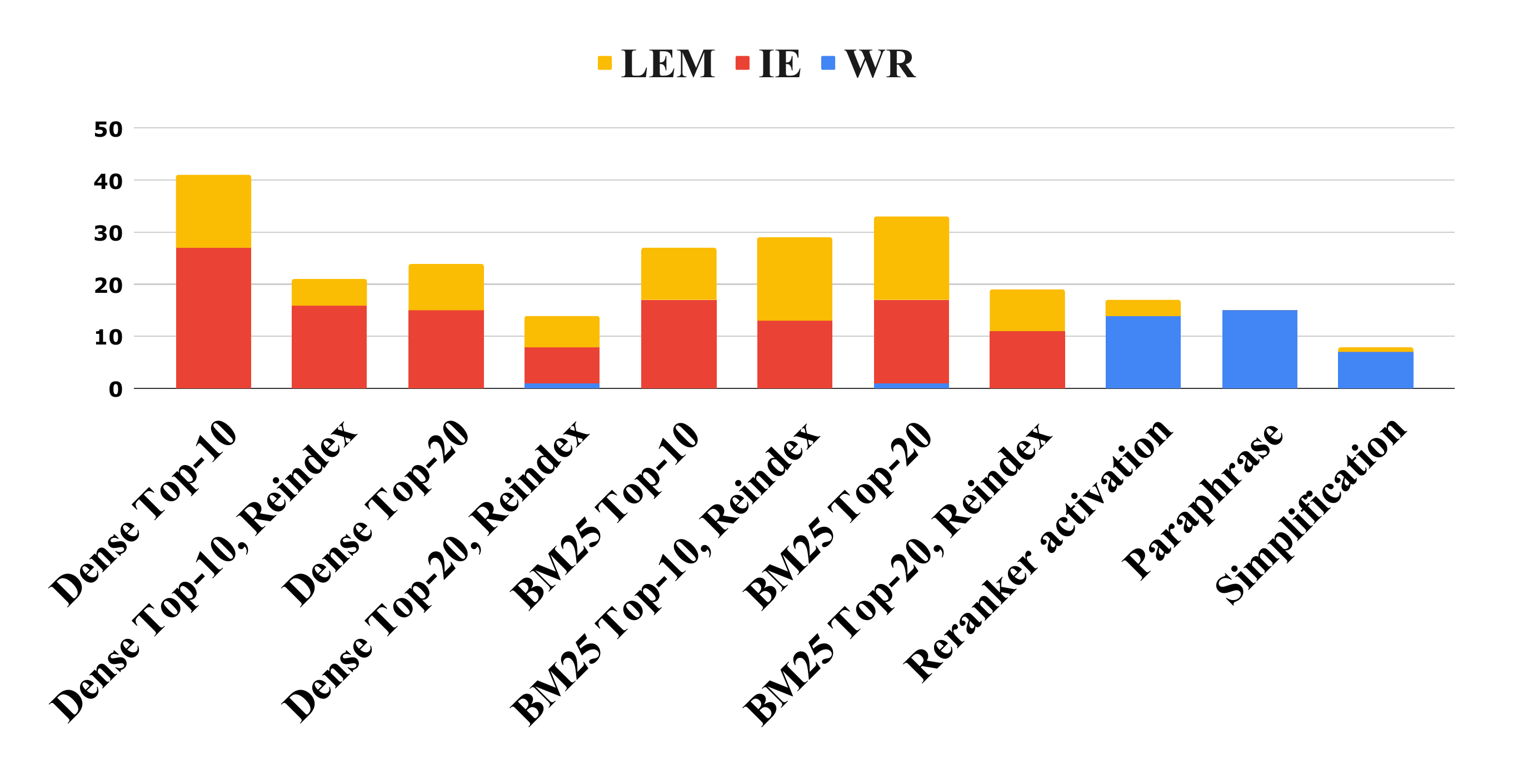}
        \caption{Stringent variant on FEVER}
    \end{subfigure}
    ~ 
    \begin{subfigure}[t]{0.5\textwidth}
        \centering
        \includegraphics[height=1.5in]{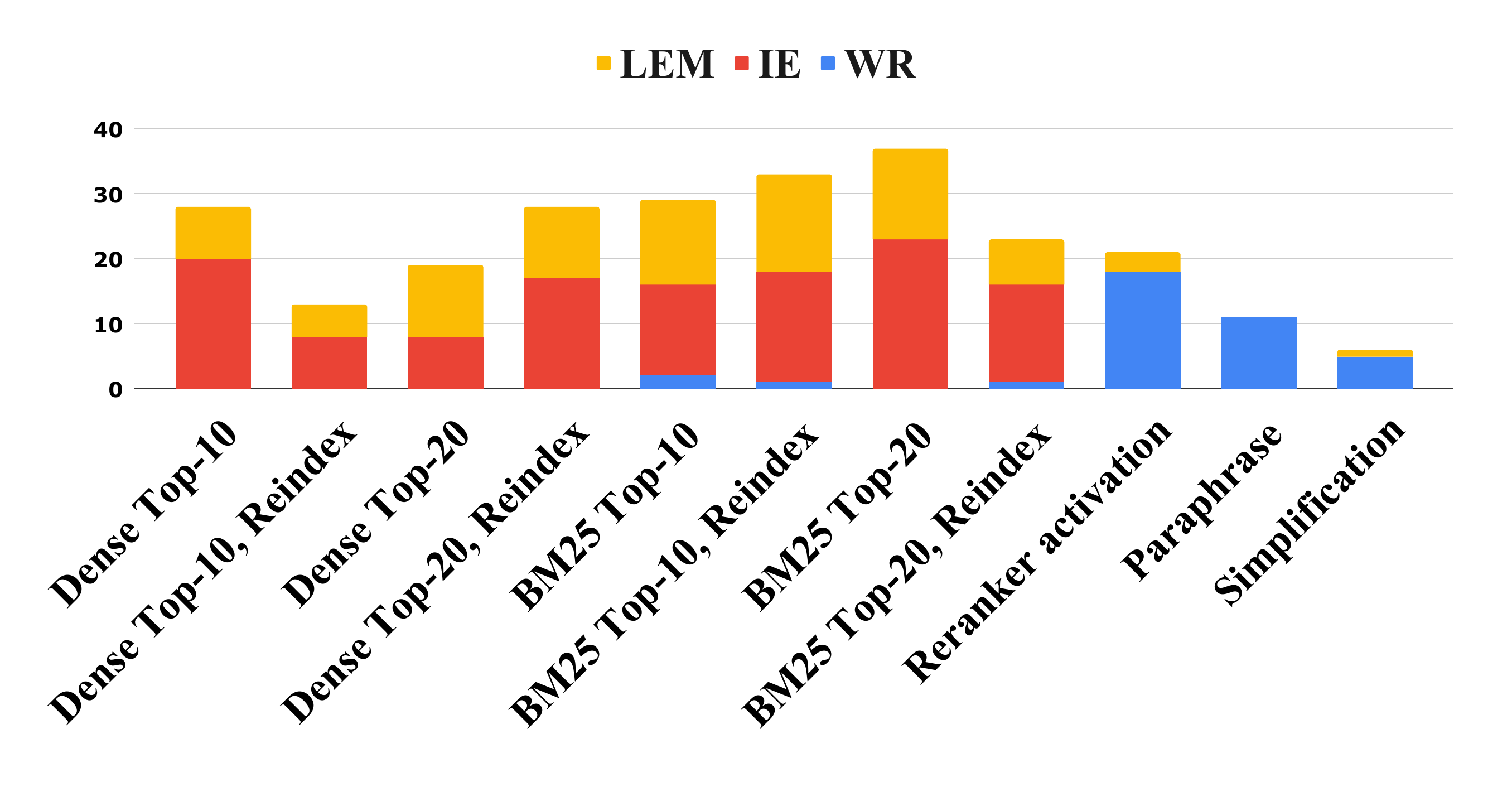}
        \caption{Relaxed variant on FEVER}
    \end{subfigure}
    \\
    \begin{subfigure}[t]{0.5\textwidth}
        \centering
        \includegraphics[height=1.5in]{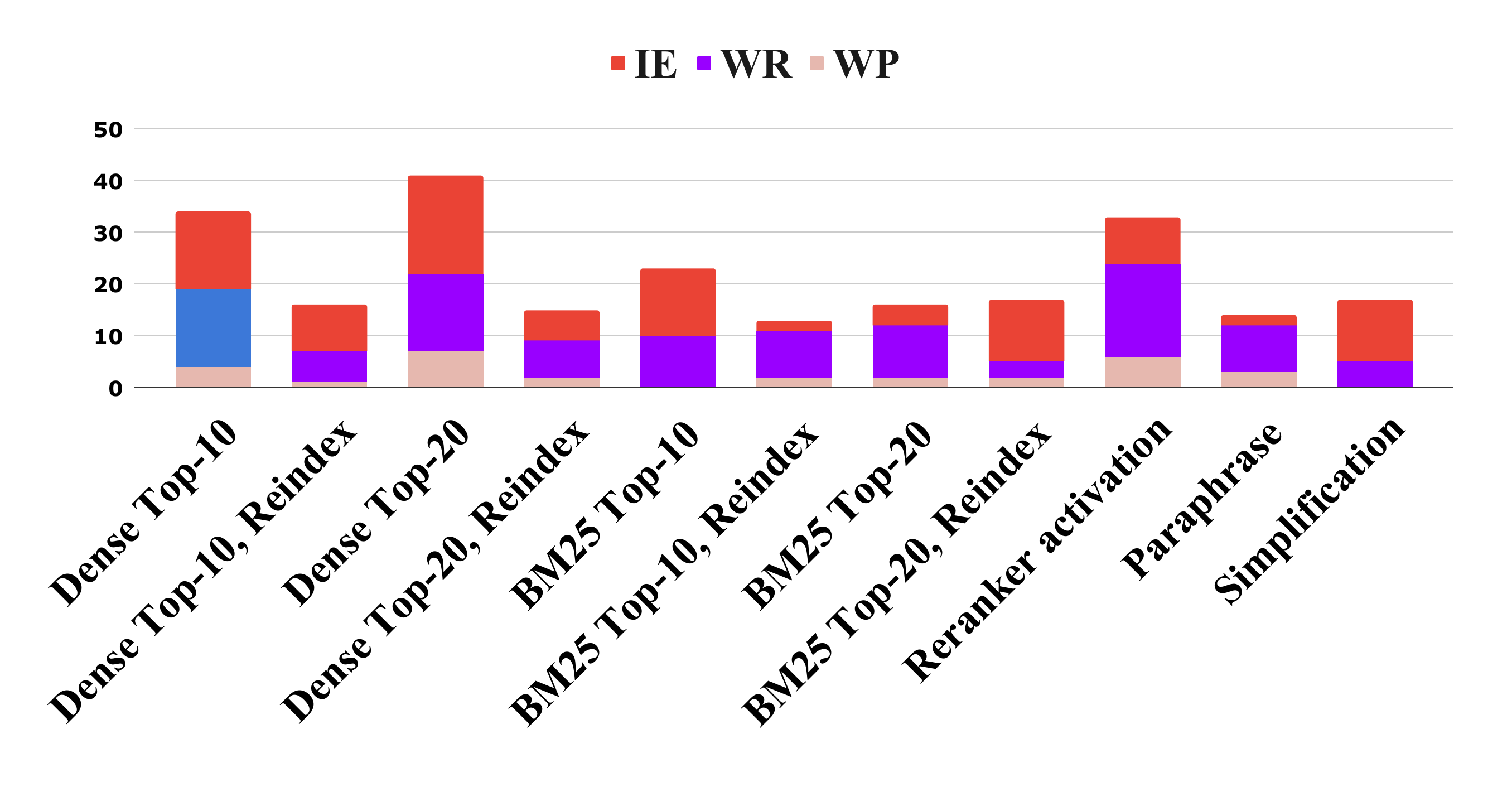}
        \caption{Stringent variant on HotpotQA}
    \end{subfigure}
    ~ 
    \begin{subfigure}[t]{0.5\textwidth}
        \centering
        \includegraphics[height=1.5in]{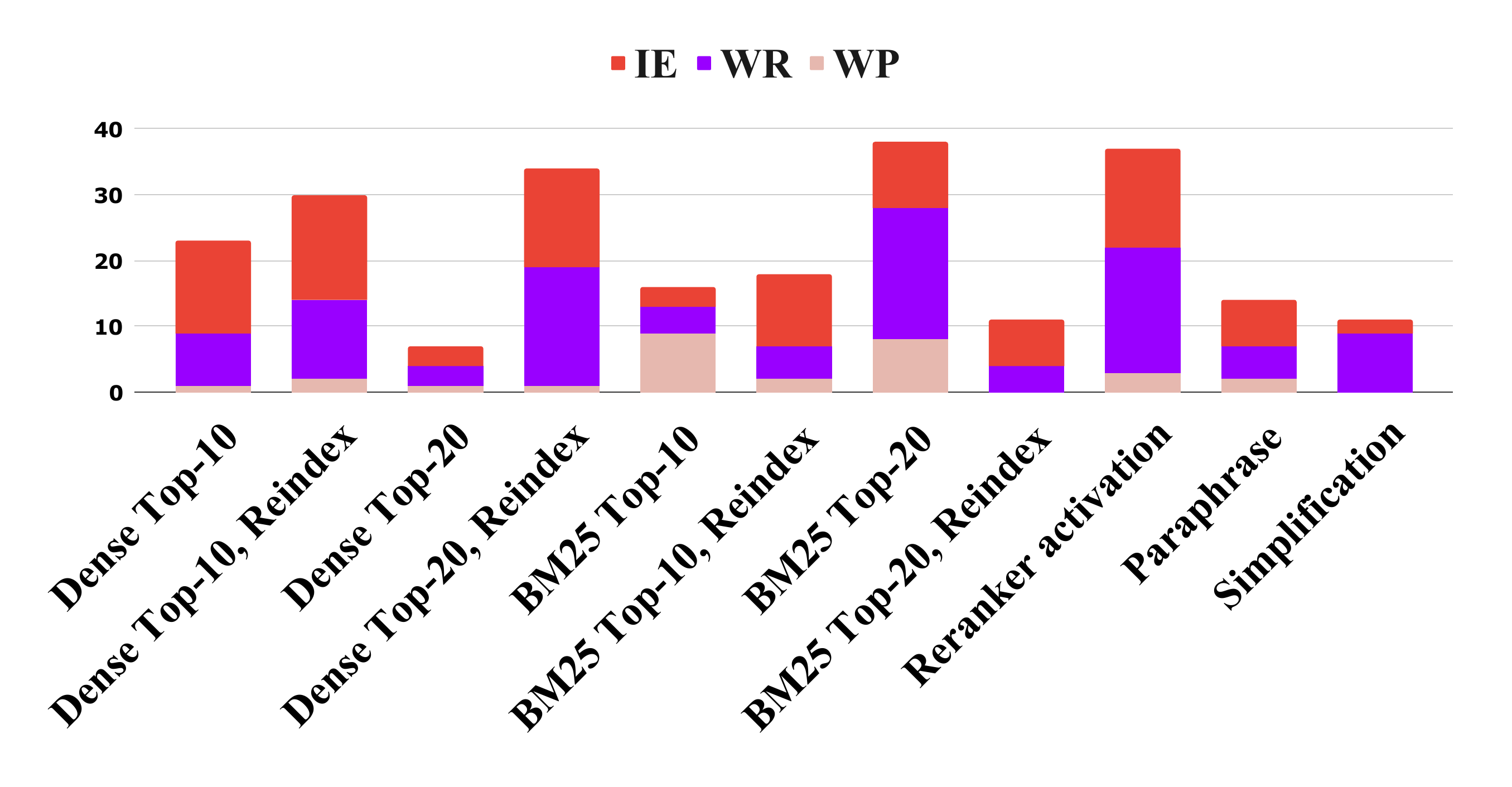}
        \caption{Relaxed variant on HotpotQA}
    \end{subfigure}
    \caption{Analysis of failure type frequency under stringent and relaxed budgets.}
    \label{fig:flabelfreq}
\end{figure}

Further budget sensitivity analysis (Figure~\ref{fig:flabelfreq}) shows that under tight budgets, the policy relies more heavily on low-cost retrieval adjustments, especially on HotpotQA.

\end{document}